\pdfoutput=1
\documentclass[letterpaper, 10 pt, conference]{ieeeconf}  

\IEEEoverridecommandlockouts                              

\usepackage{cite}
\usepackage{amsmath,amssymb,amsfonts}
\usepackage{algorithmic}
\usepackage{graphicx}
\usepackage{textcomp}
\usepackage{xcolor}
\usepackage{bm}
\usepackage{multirow}
\usepackage{subcaption}
\usepackage{etoolbox}
\usepackage{caption}
\makeatletter
\let\NAT@parse\undefined
\makeatother
\usepackage[colorlinks,
            linkcolor=black,
            anchorcolor=black,
            citecolor=black]{hyperref}
\usepackage{float}

\newcommand{\insertfig}{
    \includegraphics[width=\linewidth]{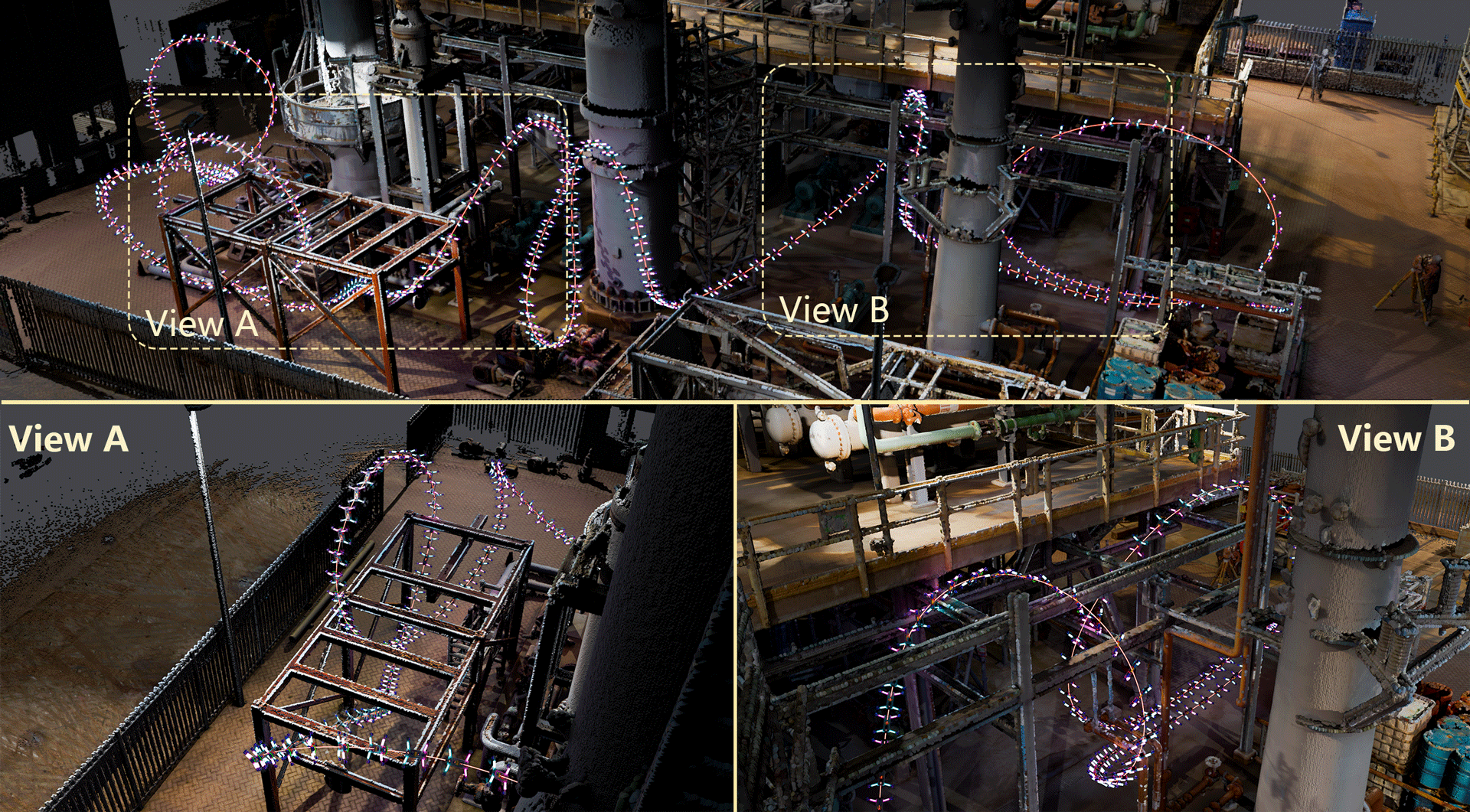}
    \captionof{figure}{Our method enables automatic generation of successive long-horizon aerobatic maneuvers, allowing drones to traverse through a complex industrial factory with dynamically feasible motion.}
    \label{fig: top}

    \vspace{-0.4cm}
    \setcounter{figure}{1}
}

\title{\LARGE \bf
Automatic Generation of Aerobatic Flight in Complex Environments\\via Diffusion Models
}

\author{Yuhang Zhong$^{1,2}$, Anke Zhao$^{1,2}$, Tianyue Wu$^{1,2}$, Tingrui Zhang$^{1,2}$ and Fei Gao$^{1,2,*}$%
\thanks{$^*$Corresponding Author: Fei Gao.}%
\thanks{$^1$Institute of Cyber-Systems and Control, College of Control Science and Engineering, Zhejiang University, Hangzhou 310027, China.}%
\thanks{$^2$Huzhou Institute of Zhejiang University, Huzhou 313000, China.}%
\thanks{E-mail:\{YuhangZhong, AnkeZhao, tianyueh8erobot, tingruizhang, fgaoaa\newline\}@zju.edu.cn}%
}

\begin{document}

\makeatletter
\apptocmd{\@maketitle}{\centering\insertfig}{}{}
\makeatother

\maketitle
\thispagestyle{empty}
\pagestyle{empty}

\begin{abstract}

Performing striking aerobatic flight in complex environments demands manual designs of key maneuvers in advance, which is intricate and time-consuming as the horizon of the trajectory performed becomes long. This paper presents a novel framework that leverages diffusion models to automate and scale up aerobatic trajectory generation. Our key innovation is the decomposition of complex maneuvers into aerobatic primitives, which are short frame sequences that act as building blocks, featuring critical aerobatic behaviors for tractable trajectory synthesis. The model learns aerobatic primitives using historical trajectory observations as dynamic priors to ensure motion continuity, with additional conditional inputs (target waypoints and optional action constraints) integrated to enable user-editable trajectory generation. During model inference, classifier guidance is incorporated with batch sampling to achieve obstacle avoidance. Additionally, the generated outcomes are refined through post-processing with spatial-temporal trajectory optimization to ensure dynamical feasibility. Extensive simulations and real-world experiments have validated the key component designs of our method, demonstrating its feasibility for deploying on real drones to achieve long-horizon aerobatic flight.


\end{abstract}

\section{INTRODUCTION}

\label{sec:Introduction}
Aerobatic freestyle flight in complex environments stands as one of the most striking and visually impressive drone-based extreme sports\cite{tezza2020let}. By planning safe but agile maneuvers and incorporating creative combinations of these highly dynamic movements, one can enable spectacular flight effects that captivate spectators alike. However, the design process can be intricate, as multiple competing requirements including obstacle avoidance, dynamic feasibility and visual impact must be simultaneously satisfied. Previous works address this problem by manually adjusting trajectory parameters such as waypoints \cite{Kaufmann2020DeepDA, Lu2022TrajectoryGA, Lu2023OnManifoldMP, Tal2022AerobaticTG} with trajectory optimization. Unfortunately, these methods remain constrained by their heavy reliance on laborious parameter tuning and domain-specific expertise in aerobatic flight, creating substantial barriers for non-expert operators attempting to design even basic aerobatic maneuvers. To bridge this gap, this paper introduces an efficient framework for the automatic generation of diverse aerobatic flights, enabling practitioners to design complex long-horizon multi-maneuver trajectories with minimal human intervention.

Recently, diffusion models have shown remarkable capacity to capture multi-modal distributions, enabling diverse generation in human motion synthesis \cite{Serifi2024RobotMD, Yi2024GeneratingHI, Cohan2024FlexibleMI}, and trajectory planning \cite{Janner2022PlanningWD, Carvalho2023MotionPD, Huang2023DiffusionbasedGO}, where models produce realistic motion sequences with diverse styles and specialized behaviors. Building on these capabilities, we explore their potential for aerobatic generation. An intuitive idea is to learn directly from long-horizon aerobatic demonstrations. However, such high-quality data remains scarce due to labor-intensive manual design and time-consuming generation processes. While some works \cite{Tseng2022EDGEED, Mishra2023GenerativeSC, Yang2023LongDanceDiffLD, Tevet2024CLoSDCT} attempt to realize long-horizon generation by combining short-horizon sequences, they face twofold critical challenges when applied in aerobatic scenarios. First, aerobatic maneuvers demand strict sequential pose transitions (e.g., continuous 360° z-axis rotation in a loop maneuver). When trajectories are fragmented into short segments for training, the model fails to capture the precise timing and coordination between sequential poses, resulting in incomplete motion generation. Second, unlike image generation \cite{Lugmayr2022RePaintIU} where pixel-level discontinuities are visually tolerable, aerobatic flight demands strict spatial-temporal continuity. Naive concatenation of short motion segments inevitably introduces visible discontinuities, which is a critical flaw given the highly dynamic nature of maneuvers requiring seamless state evolution. Consequently, it's crucial to define a modular yet kinematically consistent representation for achieving seamless composition of long-horizon aerobatic maneuvers.

In this paper, we propose to learn from aerobatic primitives to address the aforementioned challenges. An aerobatic primitive is a sequence of maneuver frames that captures key attitude changes over time and can be seamlessly combined with other primitives to achieve successive and arbitrarily long-horizon aerobatic flight. Crucially, these primitives support explicit conditioning on both maneuver styles and target waypoints, enabling user-specific trajectory generation through intuitive parameter adjustment. However, without the awareness of the previous executed primitives, the continuity of the aerobatic primitives is hard to guarantee. To mitigate this, we incorporate historical trajectory observations as transitional priors into the model architecture, allowing it to capture the latent dynamics underlying primitive transitions. While generating high-quality motions, the model is not trained with environmental information, thus providing no collision avoidance guarantees in unseen environments. We address this problem by adopting batch sampling for each primitive generation with classifier guidance \cite{Dhariwal2021DiffusionMB, Song2020ScoreBasedGM}, a widely used technique for steering the generation toward a specific target distribution. The coarse collision check on the generated trajectories is applied in each inference to further improve the obstacle avoidance success rate.

While diffusion models can generate robot-executable trajectories via positional or velocity control \cite{chi2024diffusionpolicy,Sridhar2023NoMaDGM}, they fail to meet the demands of precise control over actuator-level commands (e.g., thrust and angular velocities) during aerobatic flight. Although models implicitly encode such control signals during training, they lack explicit enforcement of dynamic feasibility, which is critical for successful flight in practical deployment. Therefore, post-processing with trajectory optimization is proposed to ensure the final trajectory stays within dynamic constraints. Notably, due to the extreme nonlinearity associated with optimizing attitude and angular velocity in the differential flatness based trajectory optimization framework \cite{Mellinger2011MinimumST, Faessler2017DifferentialFO, Wang2021RobustTP}, we design a hierarchical optimization framework to guide the final optimization to converge to a favorable local optimum, making practical deployment feasible.

Our contributions are summarized as follows:
\begin{itemize}
\item [1)]By learning from aerobatic primitives and incorporating an additional collision avoidance strategy, our diffusion model is capable of generating arbitrary long-horizon trajectories in complex environments despite being trained exclusively on short-horizon demonstration.
\item [2)]Post-processing with hierarchical trajectory optimization is designed to guarantee that generated aerobatic trajectories are physically feasible.
\item [3)]The simulation and experimental results demonstrate that the proposed method exhibits a high capability of generating a wide variety of aerobatic trajectories in complex environments.
\end{itemize}

\section{Related Works}
\label{sec:RelatedWork}
\subsection{Aerobatic Flight Generation for Quadrotors}
\label{subsec: AFG}
Generating aerobatic flight presents significant challenges due to its competing requirements for rapid attitude changes and dynamic feasibility planning. Current approaches can be broadly categorized into two paradigms. Rule-based approaches \cite{Chen2020ControllerSA, Lupashin2010ASL, Kaufmann2020DeepDA, Lu2023OnManifoldMP} employ motion decomposition strategies, in which complex maneuvers are segmented into different phases. While Kaufmann et al.\cite{Kaufmann2020DeepDA} and Lu et al.\cite{Lu2023OnManifoldMP} utilize vertical circles or arcs to enable basic 3D aerobatic motion generation beyond planar constraints, they suffer from limited adaptability to dynamic environments and require laborious parameter tuning for each specific maneuver. As the optimization method demonstrates significant success in quadrotor applications \cite{Zhou2022SwarmOM, Zhou2022RACERRC, Gao2023AdaptiveTA, Zhang2023AutoFA}, growing research formulates aerobatic generation as trajectory optimization problems to leverage their inherent flexibility. The authors in \cite{Lu2022TrajectoryGA, Tal2022AerobaticTG} achieve aerobatic trajectory generation of tail-sitter by adjusting positional- and temporal-related parameters, enabling multi-maneuver flights in open indoor and outdoor environments. However, existing methods primarily focus on aerobatic trajectory generation in open environments, neglecting essential obstacle interactions. More critically, these approaches require meticulously designed initial values to circumvent suboptimal local minima, a critical limitation stemming from the nonconvex optimization landscape created by strong nonlinearities in coupled attitude-obstacle constraints. 

\subsection{Diffusion Model for Motion Generation}
\label{subsec:dmp}
Diffusion models have emerged as a widely adopted approach for generating motions across diverse applications. In motion planning, researchers utilize diffusion models to produce trajectories characterized by optimal distributions of positions and velocities. To enforce task-specific constraints, reinforcement learning (RL) rewards \cite{Janner2022PlanningWD} or task-oriented cost functions \cite{Carvalho2023MotionPD, Huang2023DiffusionbasedGO} are integrated to guide trajectory distribution refinement. For human motion synthesis, the diffusion model’s capacity for modeling high-dimensional spaces enables learning intricate motion representations. Additional conditional inputs, such as text-guided motion styles \cite{Serifi2024RobotMD} and partial state constraints \cite{Tevet2022HumanMD} for motion generation, further enhance the editing flexibility of synthesized motions. However, the above methods primarily focus on generating fixed-length motion sequences, leaving the potential of diffusion models for long-horizon tasks underexplored. While recent studies employ policy-based methods \cite{chi2024diffusionpolicy, Sridhar2023NoMaDGM} to generate action sequences in manipulation and visual navigation tasks, their reliance on the Markov assumption often leads to myopic generation behaviors. This manifests as delayed responses to impending obstacles and fragmented execution of aerobatic maneuvers. In contrast to these approaches, our work proposes a novel framework that learns aerobatic primitives that capture key aerobatic maneuver dynamics. By strategically combining these primitives with guidance design, we achieve coherent long-horizon aerobatic motion generation while preserving consistency with physical constraints and environmental interactions.

\section{Aerobatic Diffusion Model}
\label{sec: ADM}

\subsection{Aerobatic Primitive Representation}
\label{subsec:Aerobatic Primitive}
Aerobatic primitives are expressed as a sequence of states $\bm{\tau} = \left \{ \bm{x}_{0}, \bm{x}_{1} \cdots, \bm{x}_{N_{a}}  \right \} $, $ \bm{x}_{i} = \left \{ s, \bm{p}, \bm{r} \right \} \in  \mathbb{R}^{10}$ with a constant time step, where $\bm{p} \in  \mathbb{R}^{3}$ is the position of the quadrotor and $\bm{r} \in  \mathbb{R}^{6}$ denotes a continuous 6-DoF rotation representation \cite{Zhou2018OnTC}. Notably, different maneuvers possess distinct execution durations, resulting in discrete state sequences with non-uniform lengths. This conflicts with the inherent fixed-length requirement of the diffusion model's output. To address this, a state flag $s \in \left \{ 0, 1 \right \} $ is introduced to dynamically truncate the results when s transitions from 0 to 1, where $ s = 0$ indicates the confidence that the current state belongs to the actual maneuver, while $ s = 1$ represents padding states. This simple design allows variable-length primitive generation while maintaining fixed network output dimensions. To ensure complete motion generation, we set the output sequence length $N_{a}$ to accommodate the maximum primitive duration in our dataset, with shorter sequences naturally terminated through $s$-guided truncation.

\begin{figure}[t]  
    \vspace{0.2cm}
    \centering
    \includegraphics[width=1.0\columnwidth]{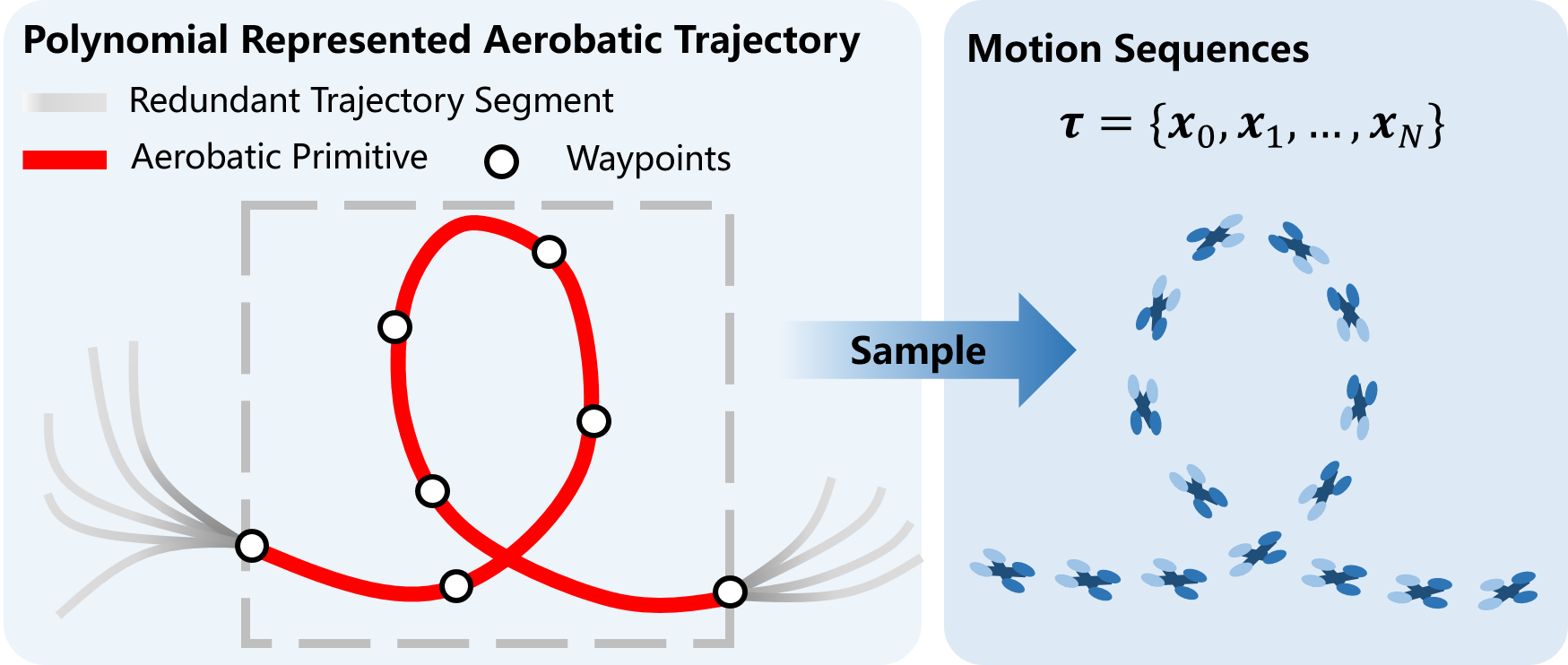}
    \caption{Illustration of aerobatic primitive generation, the trajectory segments containing the aerobatic maneuver are segmented, and the discretized motion sequences are sampled from it. Redundant trajectory segments are added to simulate the transition between aerobatic primitives.}
    \label{fig: aerobaticPrimitve}

    \vspace{-0.6cm}
\end{figure} 

\subsection{Data Preparation}
\label{subsec:Data}
We generate an expert dataset through an optimization-based method, focusing on short-horizon aerobatic maneuvers in open space. As Figure \ref{fig: aerobaticPrimitve} demonstrates, given that the generated trajectory is modeled as a continuous polynomial, aerobatic primitives are obtained by sampling from the specific segments of the complete trajectory. To facilitate dynamic transitions, we strategically prepend or append redundant trajectory segments to each primitive. This also benefits the model conditioning, as motion sequences from previous trajectory segments can serve as prior observations for learning seamless transitions. Additionally, the end state of aerobatic primitives is treated as a target waypoint and is incorporated into the model as a condition to enhance controllability. Different aerobatic maneuvers are randomly sampled based on predefined maneuver design rules. Notably, the generated demonstrations are inherently environment-agnostic by design. During model inference, we dynamically incorporate environmental context to enable obstacle-aware trajectory generation, as detailed in Section \ref{subsec:CAS}.

\subsection{Conditional Diffusion Model for Aerobatic Primitive Generation}
\label{subsec: CDM}
\begin{figure}[t]  
    \vspace{0.2cm}
    \centering
    \includegraphics[width=\linewidth]{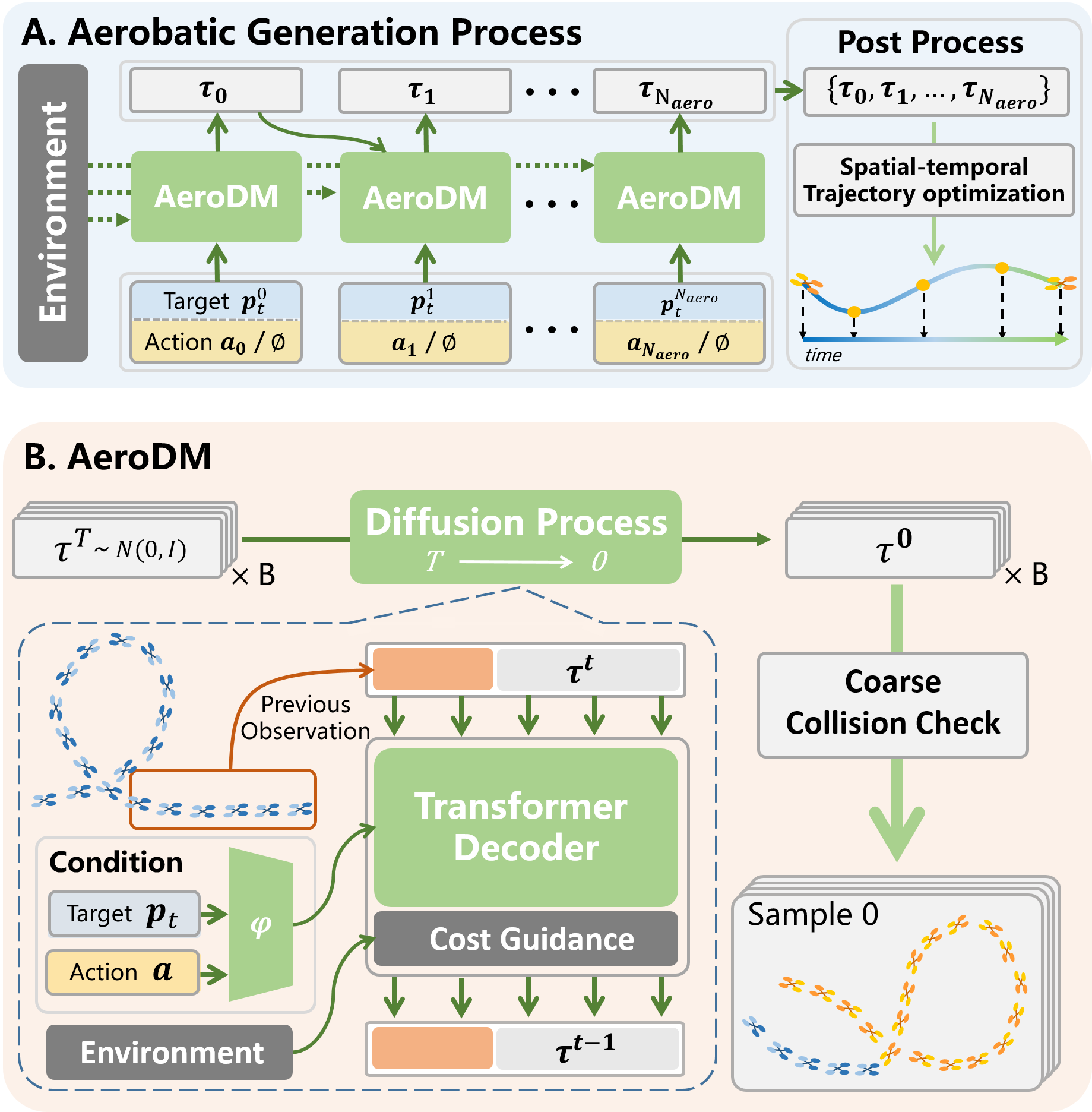}
    \caption{The architecture of the diffusion process. (A) Schematic of the overall process. 
    (B) Detailed structure of the Aerobatic Diffusion Model. 
    }
    \label{fig:network}
    \vspace{-0.6cm}
\end{figure}

We propose the Aerobatic Diffusion Model (AeroDM), a conditional diffusion model \cite{Song2020ScoreBasedGM, Dhariwal2021DiffusionMB}, to generate aerobatic primitives. This model generates the samples by learning the denoising process $p_{\theta}(\bm{\tau}^{t - 1} | \bm{\tau}^{t}, \bm{c})$ from pure Gaussian noise $\mathcal{N}(\bm{0}, \bm{I})$ to the original data distribution $p(\bm{\tau}^{0} | \bm{c})$ under the special conditions $\bm{c}$. In this paper, $\bm{c}$ contains previous state observations, target waypoint $\bm{p}_{t}$ which indicates the terminal position of $\bm{\tau}$, and action $a$ that presents the aerobatic maneuver style. The denoising process is the reverse of the forward process $q(\bm{\tau}^{t}|\bm{\tau}^{t-1}, \bm{c})$, which corrupts the data structure by gradually adding increasing noise. The predicted sample distribution can be expressed as:
\begin{align}
  p_{\theta}(\bm{\tau}^{0} | \bm{c}) = \int p(\bm{\tau}^{T}| \bm{c})\prod_{t = 1}^{T}p_{\theta}(\bm{\tau}^{t - 1}|\bm{\tau}^{t}, \bm{c})d\bm{\tau}^{1:T},
\end{align}
where $p(\bm{\tau}^{T}) = \mathcal{N}(\bm{0}, \bm{I})$. During training, the gaussian noise is added in the forward process:
\begin{align}
  q(\bm{\tau}^{t}|\bm{\tau}^{t-1}, \bm{c}) = \mathcal{N}(\sqrt{\alpha_{t}} \bm{\tau}^{t -1}, (1-\alpha_{t})I ) ,
\end{align}
where $\alpha_{t} \in (0, 1)$ are predefined scheduled parameters. Instead of predicting diffusion noise $\epsilon$, we choose to directly predict the original sample $\bm{\tau}^{0}$ to facilitate explicit geometric constraint integration. The reconstruction loss can be written as:
\begin{align}
    \mathcal{L}_{recon} = E_{\bm{\tau}_{data}\sim p(\bm{\tau}_{data}),t\sim[1,T]}[\left \| \bm{\tau}_{data} - \bm{\tau}_{\theta}(\bm{\tau}^{t}, t) \right \|^{2}_{2}] .
\end{align}
Inspired by work \cite{Tevet2022HumanMD}, we introduce velocity loss to encourage smooth transition along the primitives:
\begin{align}
    \mathcal{L}_{vel} = \frac{1}{N_{a} - 1}\sum_{i=1}^{N_{a}-1}\left \| (\bm{x}^{\theta}_{i} - \bm{x}^{\theta}_{i - 1})  -  (\bm{x}^{data}_{i} - \bm{x}^{data}_{i - 1}) \right \|^{2}_{2}  .
\end{align}
The aerobatic generation process, illustrated in Fig. \ref{fig:network}(a), operates iteratively through the aerobatic diffusion model. At each process step $i$, the model generates the current aerobatic primitive $\tau$ conditioned on target waypoints, and optional action signals with environment-related guidance. This process continues until the trajectory sequence $\left \{ ..., \bm{\tau}_{i-1},\bm{\tau}_{i} \right \} $ reaches predefined aerobatic maneuver number $N_{aero}$. After generation, the primitives are concatenated and refined in the post-processing stage. Notably, the action input can be omitted during inference to enable diverse style generation.

\subsection{Network Architecture}
\label{subsec: NA}
As illustrated in Fig. \ref{fig:network}(B), we adopt a Diffusion Transformer architecture to model the temporal dependencies in aerobatic primitive sequences. The model utilizes a decoder-only transformer backbone where both the denoising trajectory $\bm{\tau}^{t}$ (at diffusion time step $t$) and historical observations of previous primitives $\bm{\tau}^{t-1}$ are jointly processed through self-attention layers. This design explicitly enforces continuity between the generated primitive $\bm{\tau}^{t}$ and its predecessors. Additionally, conditional inputs including the denoising timestamp $t$, target waypoint $\bm{p}_{t}$, and action $a$, are first encoded via MLP embeddings ($\varphi$) separately and then integrated into the transformer through a cross-attention module.

\subsection{Collision Avoidance Strategy}
\label{subsec:CAS}
The generated aerobatic primitives have no guarantee of collision avoidance in the cluttered environment. We mitigate this problem by adding cost guidance for obstacle avoidance when doing model inference. This technique is derived from the conditional probability with Bayes Rule:
\begin{align}
    p(\bm{\tau}^{t-1} | \bm{\bm{\tau}}^{t}, O) \propto p(\bm{\tau}^{t-1} | \bm{\tau}^{t}) p(O | \bm{\tau}^{t-1}),
\end{align}
where $p(\bm{\tau}^{t-1} | \bm{\tau}^{t})$ is the denoising process, and $p(O | \bm{\tau}^{t-1})$ is the likelihood of achieving collision avoidance. By following the derivation in the work of \cite{Carvalho2023MotionPD}, the result can be approximated as Gaussian:
\begin{align}
    p(\bm{\tau}^{t-1} | \bm{\tau}^{t}, O) \approx  \mathcal{N}(\bm{\tau}^{t-1}, \mu+ \Sigma g, \Sigma),
\end{align}
where $\mu$ and $\Sigma$ are mean and variance of $p(\bm{\tau}^{t-1} | \bm{\tau}^{t})$, $g$ denotes an energy function:
\begin{align}
    \bm{g} &= \nabla_{\bm{\tau}^{t-1}} \log p(O|\bm{\tau}^{t-1})|_{\bm{\tau}^{t-1} = \bm{\mu}} \\ \nonumber
    &= \sum_{i}\lambda_{i}\nabla_{\bm{\tau}^{t-1}}c_{i}(\bm{\tau}^{t-1})|_{\bm{\tau}^{t-1} = \bm{\mu}}.
\end{align}
The collision cost function is calculated with precomputed sign distance field $sdf(\bm{x})$ from the original map, where the penalty is added when the $sdf(\bm{x})$ is smaller than d:
\begin{align}
    \bm{g_{c}}(\bm{\tau}) = \left\{\begin{matrix}
    -sdf(\bm{\tau}) + d  & sdf(\bm{\tau}) \le d \\
    0  & sdf(\bm{\tau}) >  d
    \end{matrix}\right. \ .
\end{align}
While cost guidance enhances collision avoidance rates, it cannot guarantee absolute collision-free operations. Thus, the collision probability gradually diminishes as multiple aerobatic primitives are iteratively generated from preceding ones. To address this, we implement batch sampling followed by an additional coarse collision check step for each generated outcome $\bm{\tau}$. The coarse collision check module identifies $\bm{\tau}_{i}$ violating the safety condition ( $sdf(\bm{\tau}_{i}) < 0 $ ), then iteratively modifies them by replacing colliding trajectories with randomly selected collision-free alternatives in samples. This redundant procedure effectively improves the overall success rate of aerobatic flight.

\begin{figure*}[t] 
    \vspace{0.2cm}
    \centering
    \includegraphics[width=2.03\columnwidth]{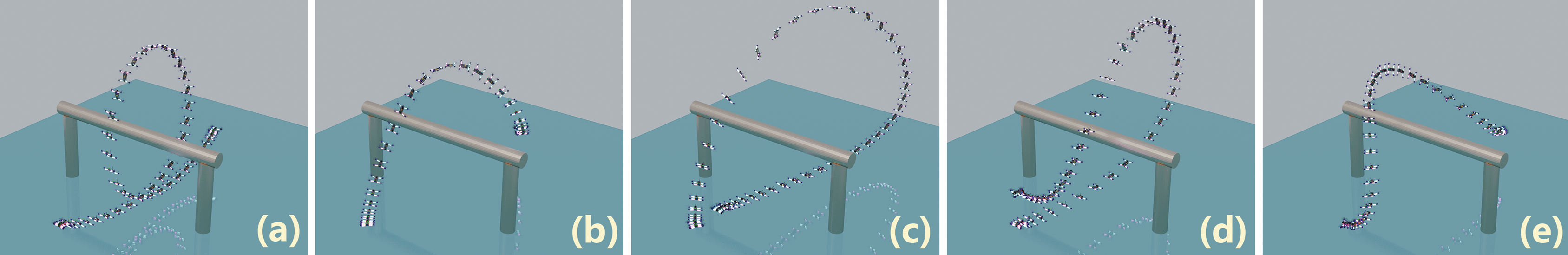}
    \caption{Five different maneuver styles of aerobatic trajectories: (a) the Power Loop, (b) the Barrel Roll, (c) the Split-S, (d) the Immelmann Turn, (e) the Wall Ride.}
    \label{fig: aerobatic}
    \vspace{-0.6cm}
\end{figure*} 

\section{Post-processing with Trajectory Optimization}
\label{sec: Trajectory Generation}

In this section, we propose to leverage spatial-temporal trajectory optimization to transform the discrete aerobatic primitives into dynamically feasible trajectories. As the section \ref{sec: ADM} suggests, the aerobatic diffusion model generates a dense sequence of frames capturing the spatial-attitude dynamics throughout the maneuvers while providing explicit topological information with collision-free properties. 

Building on this output, the waypoints presenting key attitude changes and a lightweight safety flight corridor defining free space are extracted, serving as critical inputs and constraints for trajectory optimization. We employ an iterative way to sample sparse waypoints from key frames which present the key flight maneuvers. A key frame is identified when its angular deviation in the body z-axis from its predecessor exceeds a preset threshold $\alpha$, initialized with the first frame as the reference seed. The corresponding body z-axis $\bm{z}^{ref}$ along the waypoints serves as the maneuver reference for optimization. For safe corridor generation, we utilize the method in \cite{Wang2024FastIR} to efficiently generate polyhedrons covering the whole primitive while providing sufficient free space. Additionally, the sequence generated by the diffusion model contains the optimal temporal information learned from the dataset. We directly derive inter-waypoint timestamps from it as the initial time guess for optimization.

With the aforementioned preparations, we construct the trajectory optimization problem as the following formulation:
\begin{align}
  \underset{\bm{p}(t), \bm{T}}{\min} \mathcal{J} &= \mathcal{L}_{s} + \mathcal{L}_{T} + \mathcal{L}_{att}, \\
  s.t. \ \label{eq:initx} \bm{p}^{(i)}(0) &= \bm{x}^{(i)}_{0} , \ i = 0,..., s-1, \\         
       \label{eq:finx}
        \bm{p}^{(i)}(T) &= \bm{x}^{(i)}_{f} , \ i = 0,..., s-1,  \\ 
       \mathcal{G}_{\star} &\leq 0 , \ \star = v, f_{t}, \omega \\
       \mathcal{G}_{safe} &\leq 0,
\end{align}
where the MINCO class \cite{Wang2021GeometricallyCT} is used as the trajectory representation, and the waypoints $\bm{p}(t)$ and time segment $\bm{T}$ are optimization variables. In the cost function, $\mathcal{L}_{s}$ and $ \mathcal{L}_{T}$ denote the smooth cost and time cost in normal planning problems. $\mathcal{L}_{att}$ is the cost to align the flight maneuver with the key frame reference as it is expressed as:
\begin{align}
    \mathcal{L}_{att} &= \sum_{i=0}^{n} -\cos(\frac{\bm{f}_{t}(T_{s}(i))^{\top} \bm{z}^{ref}_{i}}{\left \| \bm{f}_{t}(T_{s}(i)) \right \| } ),
\end{align}
where $T_{s}(i) = \sum_{j = 0}^{i} T_{j}$, and $\bm{f}_{t}$ denotes the net thrust in the world frame, which can be calculated based on differential flatness. Kinodynamic constraints $\mathcal{G}_{\star}$ are introduced on velocity $\bm{v}$, net thrust $\bm{f}_{t}$ and angular velocity $\bm{\omega}$, detailed in work \cite{Wang2021RobustTP}. To ensure safety during flight, we constrain each trajectory segment must stay inside the corresponding $i$th polyhedron:
\begin{align}
  \mathcal{G}_{safe} = \int_{0}^{T_{sum}}\bm{A}_{i}\bm{p}{(t)} - \bm{b}_{i} \ dt,
\end{align}
where $\bm{A}_{i}$ and $\bm{b}_{i}$ are corresponding parameters of polyhedrons. During the optimization process, we observed that the strong nonlinearity of the z-axis angular velocity can cause the optimization to get trapped in bad suboptimal local minima. This leads to significant violations of angular velocity constraints, which in turn negatively impacts the actual flight performance. To address this issue, we propose a hierarchical optimization strategy that operates in two sequential stages. First, we solve a relaxed problem formulation by temporarily removing z-axis angular velocity constraints to circumvent local minima. This initial solution then serves as a warm start for the second stage, where we perform a fully constrained optimization refinement that reinstates all dynamic constraints. This simple design improves overall optimization performance while maintaining strict dynamic constraints, thereby ensuring stunning flight performance as detailed in Sec. \ref{subsec: Real-World Experiment}.

\section{Results}
In this section, we present a series of experiments to validate the key component designs of our method and evaluate the performance in real-world scenarios. We demonstrate that
\begin{itemize}
    \item[1)] Explicit conditioning on target points and action semantics enables improved editability of generated aerobatic trajectories.
    \item[2)] Historical state integration mitigates abrupt transitions between motion primitives while preserving agility.
    \item[3)] The proposed collision avoidance strategy significantly improves the success rate of aerobatic flight in cluttered environments.
    \item[4)]The post-processing is essential for bridging discrete planning to dynamically executable trajectories in real-world deployment.
\end{itemize}

\subsection{Implementation details}
\label{subsec: Implementation details}
Our model is trained on a dataset comprising five distinct aerobatic primitives (Fig. \ref{fig: aerobatic}), where each primitive is generated by uniformly sampling target waypoints within the spatial bounds of $\left [ 0, 8 \right ] \times \left [ -6, 6 \right ]\times [-1,1] $ at a resolution of 1.0 meter. To model dynamic transitions between aerobatic primitives, the primitives are generated with supplementary trajectories by randomly sampling waypoints before and after each primitive. The dataset is further augmented through transformations in the global coordinate system, specifically applying discrete z-axis rotations of {90°, 180°, 270°} to each of the aerobatic primitives. This symmetry utilization enables omnidirectional maneuver generation while preserving dynamic feasibility constraints, ultimately yielding 450,000 training primitives. The network architecture employs a decoder-only transformer with 4 layers, 4 multi-head attention, and a latent dimension of 256. For the diffusion process, we configure 30 denoising steps with an exponential-noise scheduler. Each primitive sequence spans 6 seconds, discretized into $N_{a} = 60$ time steps at 0.1 s intervals. To ensure transition continuity, the model incorporates 5-frame historical observations as prior context.

\subsection{Simulation Ablations }
\label{subsec: Simulation Experiment}
\subsubsection{Target and Action Conditions}

\begin{figure}[t]
  \vspace{0.2cm}
  \centering
  \begin{subfigure}[b]{\linewidth}
    \includegraphics[width=\textwidth]{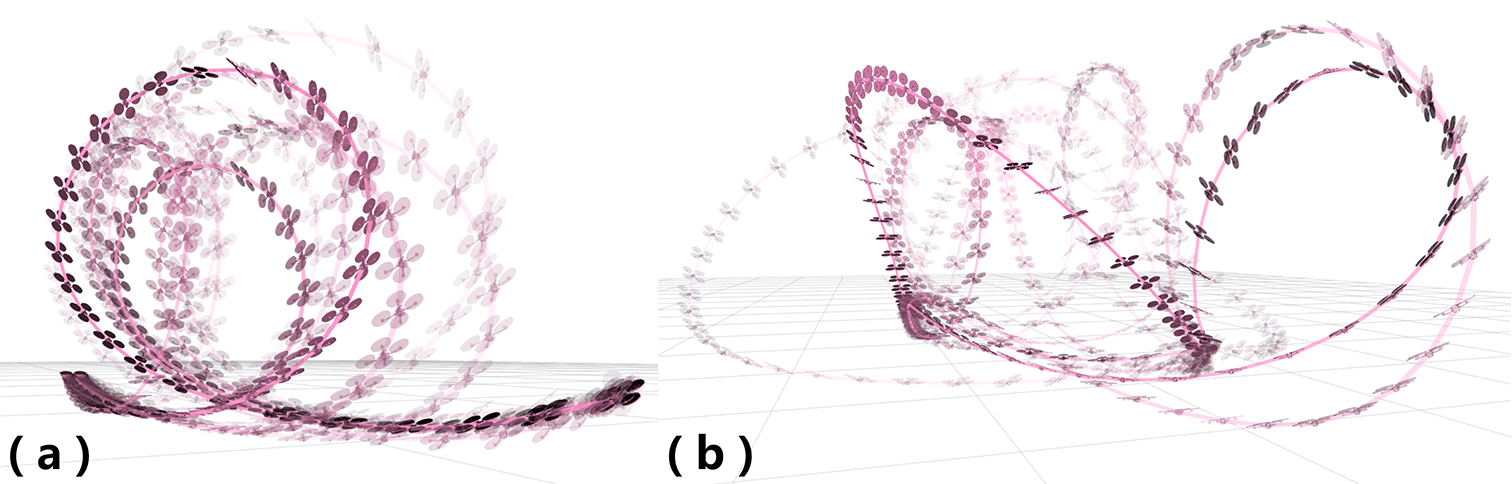}
    \label{fig:sub1}
  \end{subfigure}

  \vspace{0.1cm}

  \begin{subfigure}[b]{\linewidth}
    \includegraphics[width=\textwidth]{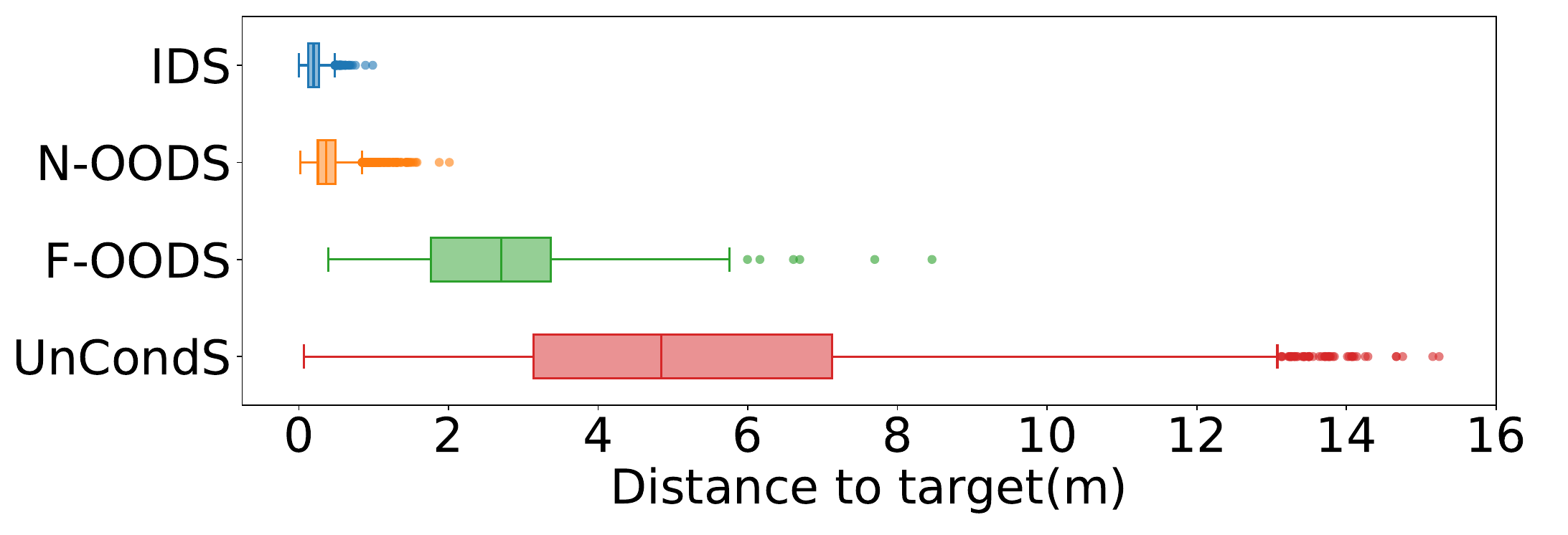}
    \label{fig:sub2}
  \end{subfigure}

  \vspace{-0.2cm}

  \caption{Up: results of the aerobatic generation conditioned on the ``Power Loop" action (a) compared with action-agnostic model (b). Down: box plot of the errors between the terminal of aerobatic primitives and the given target, measured by distances.}
  \label{fig:target action}
  \vspace{-0.4cm}
\end{figure}

To validate the effectiveness of target and action conditioning, we compare three model variants: unconstrained generation, target-only conditioning, and target-action joint conditioning. To evaluate target conditioning, 5,000 aerobatic primitives are generated with 
 50 randomly sampled target waypoints in each of the following four scenarios:
\begin{itemize}
\item[$\bullet$]
\textbf{In-Distribution Sampling (IDS)}: Targets sampled within the training distribution.
\end{itemize}

\begin{itemize}
\item[$\bullet$]
\textbf{Near Out-of-Distribution Sampling (N-OODS)}: Targets sampled outside but near the distribution boundary, defined as $\left[ 9, 12 \right ] \times \left [ -9, 9 \right ]\times [-1,1] $.
\end{itemize}

\begin{itemize}
\item[$\bullet$]
\textbf{Far Out-of-Distribution Sampling (F-OODS)}: Targets sampled far outside the distribution, defined as $\left[ 12, 16 \right ] \times \left [ -12, 12 \right ]\times [-1,1] $.
\end{itemize}

\begin{itemize}
\item[$\bullet$]
\textbf{Unconditional Sampling (UncondS)}: Targets within the distribution but generated by the unconstrained model.
\end{itemize}
The distribution of distances between primitive terminal positions and target waypoints is visualized as box plots in Fig. \ref{fig:target action}. Our analysis reveals that the target-only conditioning model reliably guides trajectories to terminate near specified targets, whereas the unconditioned one scatters endpoints randomly due to the absence of target awareness. Surprisingly, the model generalizes to targets near the distribution boundary, demonstrating generalization ability. 

To evaluate action conditioning, Fig. \ref{fig:target action} visualizes 10 trajectories generated by the target-only and target-action joint conditioning models with the same target. When given ``Power Loop" commands, the action-conditioned model produces maneuvers explicitly aligned with semantic intent (Fig. \ref{fig:target action}(a)), while the action-agnostic model generates inconsistent maneuvers. The above ablation results suggest that target conditioning provides explicit spatial guidance, while action conditioning allows flexible trajectory shaping through human-defined commands, enabling controllable generation of task-oriented aerobatic maneuvers.

\subsubsection{Transition Smoothness}
\begin{figure}[t]  
    \vspace{0.2cm}
    \centering
    \includegraphics[width=1.0\columnwidth]{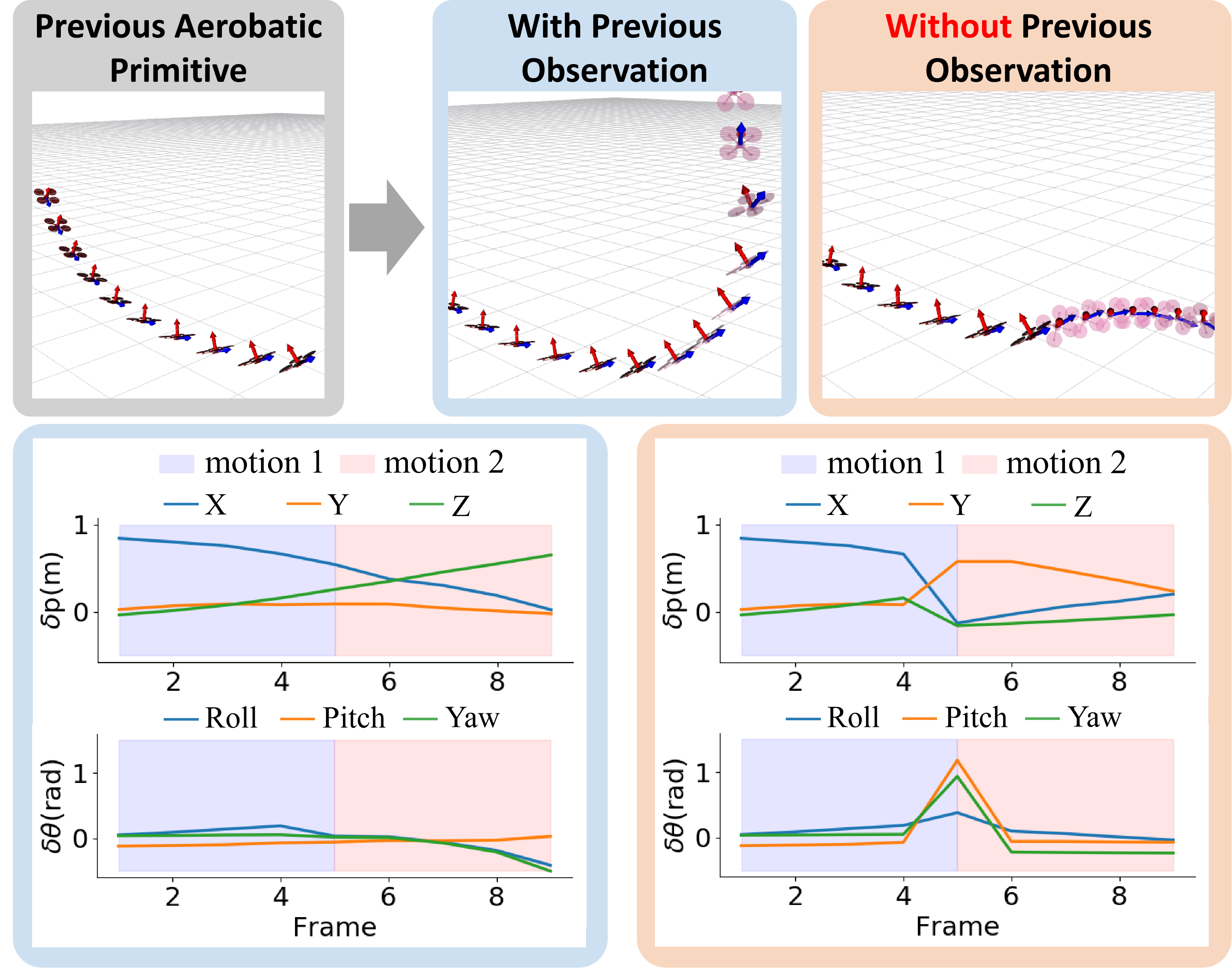}
    \caption{Comparison between models with and without access to previous primitives. The smoothness is measured with differences of positions $\delta p$ and Euler angles $\delta \theta$ between adjacent frames.}
    \label{fig: transitions}
    \vspace{-0.5cm}
\end{figure}

To validate the necessity of historical state conditioning for smooth dynamic transitions, we compare two model variants: with and without access to previous primitive states. Both models are tasked to generate the aerobatic primitives from the same previous trajectory.  We quantify motion smoothness by computing adjacent-frame differences in axis-aligned position $\delta p$ and Euler angles $\delta \theta$. The results are shown in Fig. \ref{fig: transitions}, the model without previous observations failed to understand the dynamic transition process, resulting in abrupt changes in attitude (more than 1 rad) and position (more than 0.5 m) that are infeasible in actual flight. In contrast, the model with observations generates state-coherent maneuvers even during aggressive transitions. Notably, it achieves smooth attitude adjustments and continuous positional updates that align with drone dynamics. This suggests that our model successfully learns the underlying dynamics of drone flight.

\begin{figure*}[t]  
    \vspace{0.2cm}
    \centering
    \includegraphics[width=2.0\columnwidth]{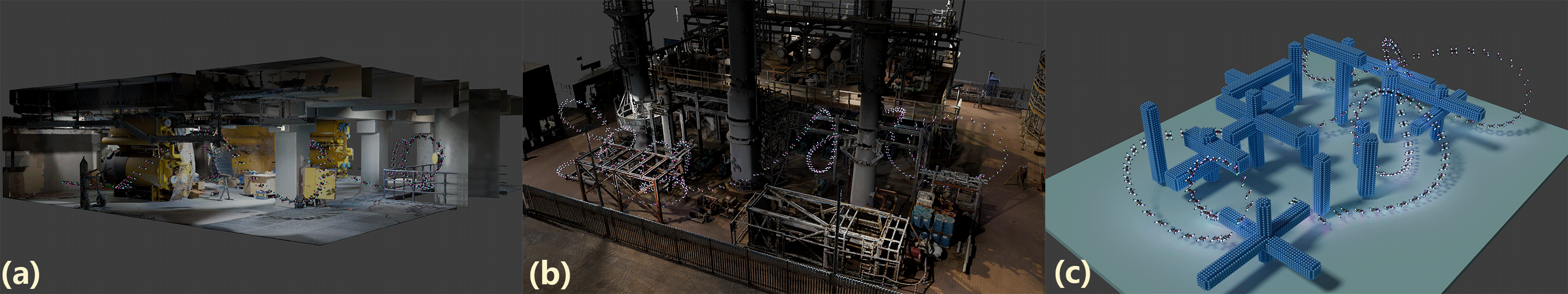}
    \caption{The illustration of the drone executing aerobatic maneuvers in three different scenarios: (a) Narrow indoor industrial workshop, (b) Complex outdoor industrial factory, and (c) random forest.}
    \label{fig: scenarios} 
\end{figure*}

\renewcommand{\arraystretch}{1.1}
\begin{table*}[t]
\centering
\caption{Success rate of different ablation configurations across environments.}
\label{table: collision}
\begin{tabular}{ccccccc}
\hline
\multicolumn{1}{l}{}             & $N_{aero}$        & 1(\%)               & 2(\%)              & 3(\%)             & 5(\%)             & 10(\%)            \\ \hline
\multirow{3}{*}{Random Forest}   & Ours     & \bm{$99.9 \pm 0.1$} & \bm{$99.8 \pm 0.2$}  & \bm{$99.7 \pm 0.3$} & \bm{$99.7 \pm 0.3$} & \bm{$99.4 \pm 0.6$} \\
                                 & UnGuided & $51.8 \pm 8.0$            & $7.0 \pm 3.0$            & $3.1 \pm 2.1$           & $0.7 \pm 0.7$           & $0.0 \pm 0.0$           \\
                                 & UnCheck  & $97.7 \pm 2.1$            & $85.4 \pm 4.2$           & $79.9 \pm 5.7$          & $72.0 \pm 6.0$          & $53.6 \pm 9.4$          \\ \hline
\multirow{3}{*}{Outdoor Factory} & Ours     & \bm{$100.0 \pm 0.0$}  & \bm{$99.9 \pm 0.1$}  & \bm{$99.8 \pm 0.2$} & \bm{$99.5 \pm 0.5$} & \bm{$97.2 \pm 2.8$} \\
                                 & UnGuided & $57.5 \pm 3.5$            & $9.7 \pm 0.9$            & $5.8 \pm 1.2$           & $0.0 \pm 0.0$           & $0.0 \pm 0.0$           \\
                                 & UnCheck  & $97.0 \pm 1.0$            & $82.5 \pm 1.7$           & $61.9 \pm 3.3$          & $23.9 \pm 3.3$          & $7.0 \pm 1.6$           \\ \hline
\multirow{3}{*}{Indoor Workshop} & Ours     & \bm{$99.9 \pm 0.1$}   & \bm{$100.0 \pm 0.0$} & \bm{$97.9 \pm 2.1$} & \bm{$98.4 \pm 1.6$} & \bm{$97.1 \pm 2.9$} \\
                                 & UnGuided & $65.7 \pm 18.1$           & $43.3 \pm 20.3$          & $15.4 \pm 9.8$          & $10.7 \pm 8.5$          & $0.4 \pm 0.4$           \\
                                 & UnCheck  & $81.1 \pm 12.3$           & $62.9 \pm 19.9$          & $35.6 \pm 17.0$         & $26. \pm 16.5$          & $2.6 \pm 2.6$           \\ \hline
\end{tabular}

\vspace{-0.5cm}
\end{table*}

\subsubsection{Collision Avoidance}
In this task, we test our method in three different scenarios illustrated in Fig. \ref{fig: scenarios}. The factory environment contains small, complex, and unstructured obstacles, while the indoor industrial workshop is extremely narrow with walls obstructing the space, presenting significant challenges for our collision avoidance strategy. $\bm{p}_t$ are set to be collision-free to guide the aerobatic generation traversing the complex environment and covering the whole flying region. To obtain reliable results, we conduct ablation tests on five different random seeds, where the batch size for sampling in each inference is set to 500. In each ablation, we compare our method with an \textbf{UnGuided} baseline (generating samples without cost guidance) and an \textbf{UnCheck} baseline (no coarse collision check implemented). The success rate is measured by the proportion of collision-free trajectories among all generated trajectories, where more precise collision checks are performed on both individual motion frames and interpolated trajectories between consecutive frames. For each baseline, the median success rate and its fluctuation range across all seeds are statistically evaluated as the number of aerobatic maneuvers $N_{aero}$ increases. 

Experimental results are summarized in Table \ref{table: collision}. Our method demonstrates superior success rates across all scenarios compared to two baselines. The comparison reveals that cost guidance contributes most significantly to collision avoidance but cannot guarantee collision-free trajectories in all cases. Specifically, trajectories generated from previously collided motion primitives may compromise subsequent collision-free generation. To address this limitation, the coarse collision check module serves as a lightweight yet effective safeguard, intercepting collision risks in the final trajectory generation process.

\begin{figure}[t]
    \centering
    \includegraphics[width=0.95\linewidth]{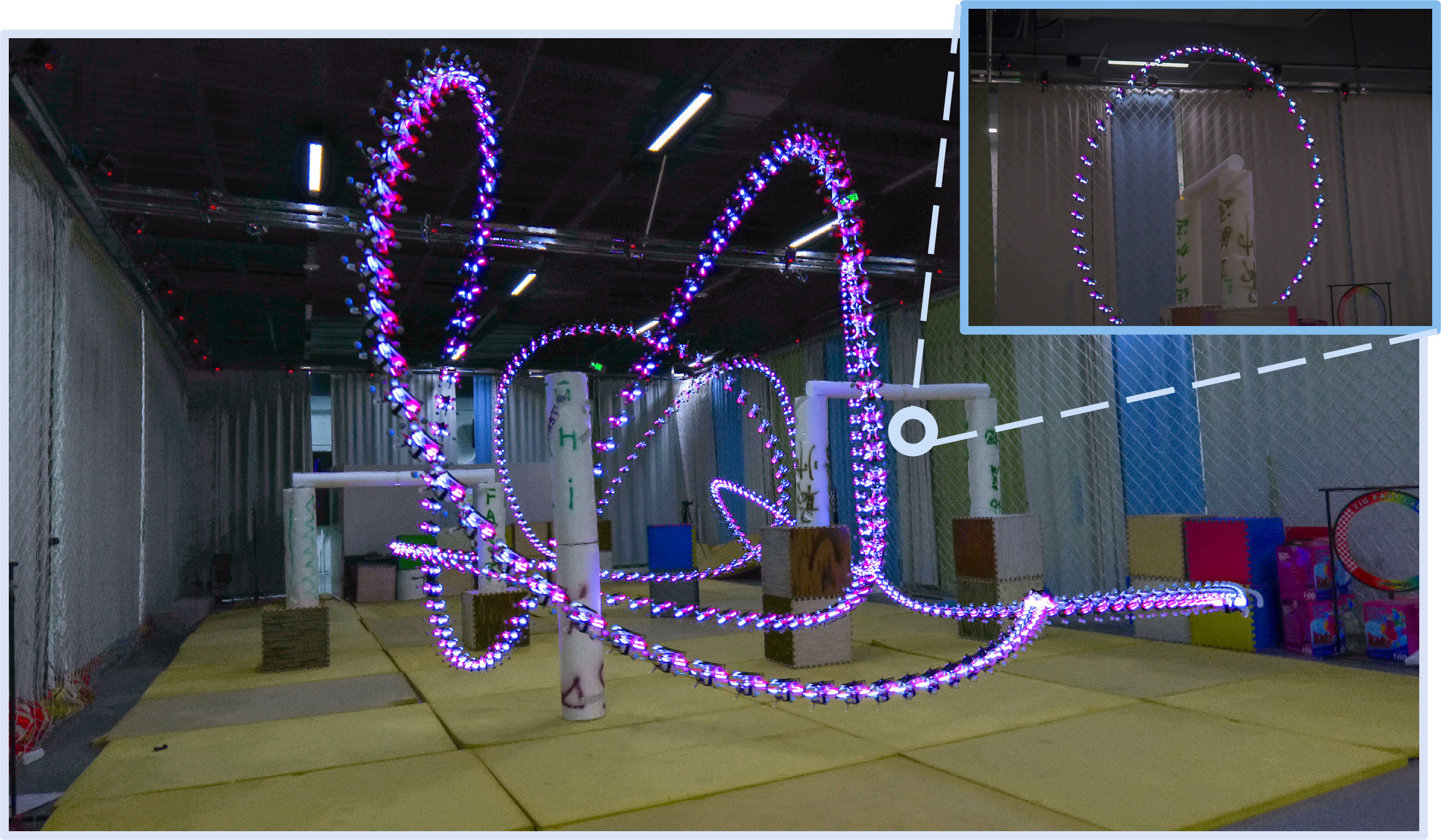}
    \caption{Snapshot of a quadrotor executing aerobatic flight trajectories with five distinct maneuvers generated by the proposed method in real-world.}
    \label{fig:realexp}
    \vspace{-0.5cm}
\end{figure}

\begin{figure}[t]  
\centering
{\includegraphics[width=1.0\columnwidth]{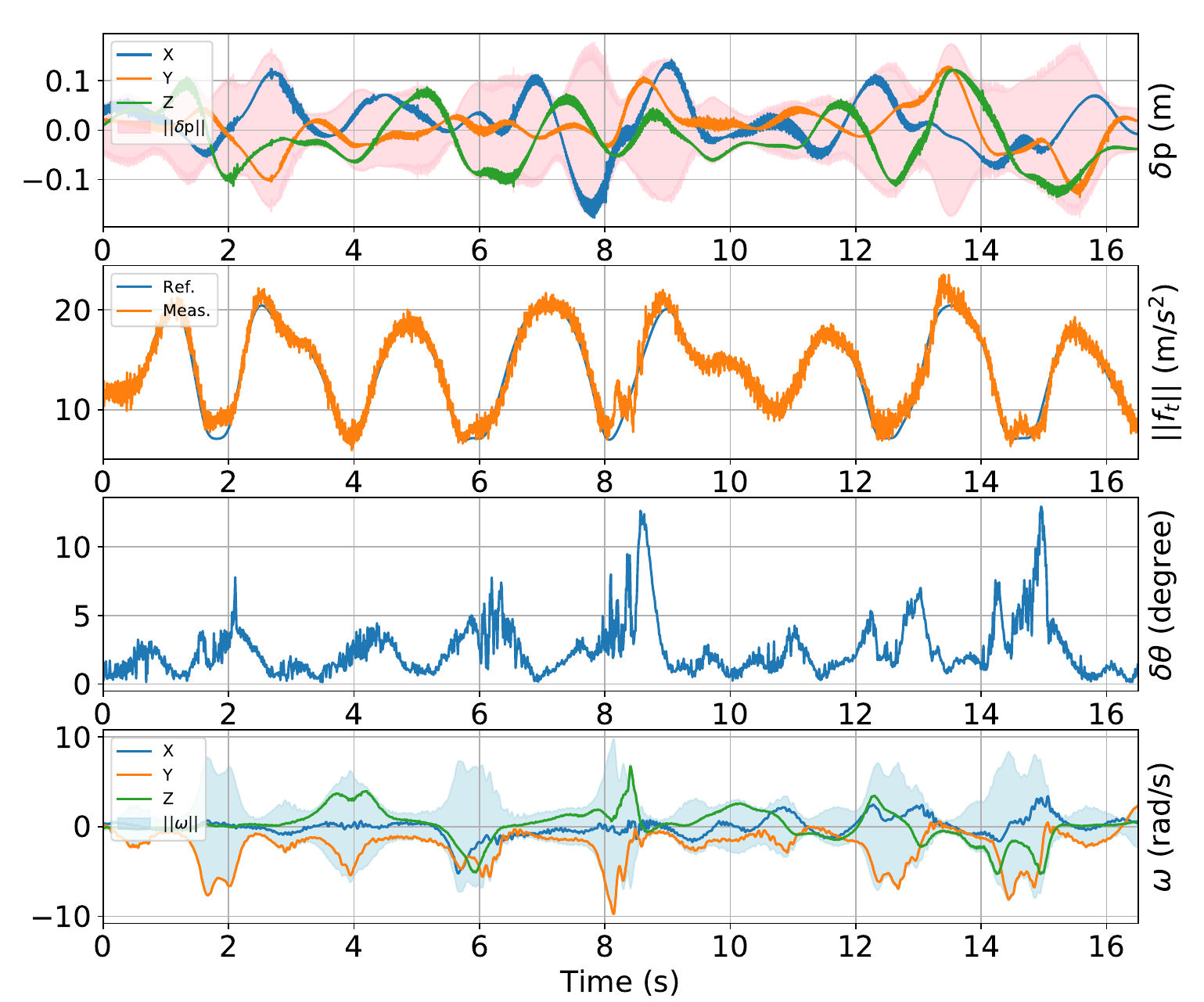}}
\caption{ \label{fig: flight data} Numerical analysis, $\delta p$ demonstrates the position error along axis $X,Y,Z$ and total tracking error $\left \| \delta p \right \| $. Ref. and Meas. denotes the net thrust values $\left \| \bm{f}_{t} \right \|$ obtained from planned trajectory and the practical measured value calculated from an inertial measurement unit (IMU) respectively. $\delta \theta$ indicates the rotational angle corresponding to the quaternion that describes the error between desired and actual attitude. $\omega$ is the angular velocity along the flight as angular velocity $X,Y,Z$ along the axis and norm $\left \| \omega \right \|$. }
\vspace{-0.5cm}
\end{figure}

\subsection{The Real-World Experiment }
\label{subsec: Real-World Experiment}
To verify the real-world applicability of the proposed method, the aerobatic trajectories containing five aerobatic maneuvers are generated in a narrow and cluttered indoor space with the size of $12 \times 6\times4\ m^3$, where a drone executes these trajectories under the NOKOV Motion Capture System\footnote{https://www.nokov.com/}. The real-world performance is demonstrated in Fig. \ref{fig:realexp}, with the numerical analysis provided in Fig. \ref{fig: flight data}. As Fig. \ref{fig: flight data} indicates, the post-processing constrains the thrust and angular velocity to remain within feasible values, ensuring that the low-level controllers can accurately track the control signals. As a result, the tracking errors in both attitude and position are small, with maximum errors of less than 15 degrees and 0.15 meter respectively. This highlights the critical role of post-processing in practical aerobatic flight generation. Since the discrete outputs of positional and attitude references from the diffusion model would be challenging for the controller to precisely track at the actuator-level commands, they would lead to potential flight failures.

\section{Conclusion and Future Work}
\label{sec:Conclusion}
In this work, we unlock the potential of diffusion models for generating long-horizon, multi-maneuver aerobatic trajectories. Our key contribution lies in learning aerobatic primitives with specific conditioning and guidance from trajectory costs, enabling automatic and editable generation. The post-processing further ensures dynamic feasibility, making the method directly deployable on physical drones in the real world. However, the proposed method achieves interaction with the environment primarily through passive obstacle avoidance, which limits the ability to generate truly visually impressive maneuvers that reflect a deep understanding of the environment. Therefore, future work will focus on developing scene-aware aerobatic generation, where trajectories are dynamically crafted in response to environmental features (e.g., flips through narrow gaps). With a better understanding of the environments, we believe that it can create more visually appealing maneuvers, blending agility with surroundings in a way that enhances both performance and aesthetic value.






\bibliographystyle{ieeetr}
\bibliography{main}

\end{document}